\documentclass{iet-ell}
\usepackage{xcolor}
\usepackage{algorithmic}
\usepackage{algorithm}
\usepackage{bbding}
\usepackage{pifont}
\usepackage{wasysym}

\setcopyright{open}
\ietVolume{60}
\ietYear{2024}
\ietDoi{ell2.13191}

\received{23 January 2024}
\accepted{27 March 2024}

\begin{document}
\newcommand{\red}{\textcolor{red}}
\newcommand{\green}{\textcolor{green}}
\title{DeepKalPose:  An Enhanced Deep-Learning Kalman Filter for Temporally Consistent Monocular Vehicle Pose Estimation}

\author[af1,af2]{Leandro Di Bella}
\orcid{0009-0000-1731-7205}
\author[af1,af2]{Yangxintong Lyu}
\orcid{0000-0002-2501-9010}
\author[af1,af2]{Adrian Munteanu}
\orcid{0000-0001-7290-0428}

\affil[af1]{Department of Electronics and Informatics, Vrije Universiteit Brussel, B-1050 Brussels, Belgium}
\affil[af2]{imec, Kapeldreef 75, B-3001 Leuven, Belgium}
\corresp{Email: yangxintong.lyu@vub.be}

\begin{abstract}
This paper presents {DeepKalPose}, a novel approach for enhancing temporal consistency in monocular vehicle pose estimation applied on video through a deep-learning-based Kalman Filter. By integrating a Bi-directional Kalman filter strategy utilizing forward and backward time-series processing, combined with a learnable motion model to represent complex motion patterns, our method significantly improves pose accuracy and robustness across various conditions, particularly for occluded or distant vehicles. Experimental validation on the KITTI dataset confirms that {DeepKalPose} outperforms existing methods in both pose accuracy and temporal consistency.
\end{abstract}

\maketitle
\vspace{-0.3cm}
\section{Introduction}
In recent years, the importance of scene understanding has become increasingly important, particularly in the development of technologies for smart mobility and intelligent transportation systems. The need to comprehend dynamic urban environments underscores the growing need for accurate video-based vehicle 6D pose estimation as real-world environments inherently give rise to dynamic data, such as video streams. However, traditional approaches \cite{lyu2022mono6d, Ke2020GSNetJV, wu_6d-vnet_2019, inproceedings} in this domain have primarily focused on image-based methodologies, which often yield temporal pose inconsistencies when applied to video data, processing each frame independently. These inconsistencies manifest as flickering artifacts, where vehicles exhibit jittery or unstable poses across successive frames. This phenomenon is primarily attributed to the absence of temporal constraints and coherency in the estimation process, significantly affecting scenarios involving occluded or distant vehicles leading to huge differences in predicted pose. This means they do not account for the continuity between frames in a video. To address this challenge, a range of innovative methods have been developed in the field of vehicle pose estimation that typically incorporate mechanisms to ensure temporal consistency, thereby reducing such artifacts.

A large body of research \cite{6599106, Giancola2019LeveragingSC, article_Shape}  has focused on shape completion where candidate shapes are encoded and compared against a predicted model shape, imposing temporal constraint. Yet, these methods often depend on additional data such as CAD models or dense point clouds, which are not always available. More recent works \cite{9760217, Hu2018JointM3, marinello2022triplettrack} propose to impose temporal relationship between frames by leveraging an LSTM-based model for motion learning. Nevertheless, LSTM approaches can suffer from challenges related to far-away detection as well as long-term dependencies. In contrast, particle filter-based methods, like Poisson multi-Bernoulli mixture (PMBM) tracking filter \cite{article_pmbm}, have demonstrated robustness against such challenges. Moreover, \cite{inbook, 9626850, 9939165, wang2023camo} integrate a 3D Kalman Filter (KF) into their frameworks, leveraging the kinematic motion of vehicles for smoothing and tracking tasks. However, challenges arise when dealing with systems where the underlying dynamics are unknown or have highly nonlinear behavior. Under these conditions, creating a mathematical model for motion becomes very complex.

In this letter, we propose an enhanced deep-learning Kalman Filter-based method (EDLFK) for temporally consistent vehicle pose estimation, dubbed \textit{DeepKalPose} which is adept at reinforcing temporal consistency that overcomes the mentioned drawbacks. It can effectively address flickering artifacts in vehicle pose estimation by providing more stable and consistent tracking of the vehicle's pose. Our method advances beyond the standard KF by incorporating a bi-branch network for forward and backward time-series analysis. Additionally, our filter integrates deep learning techniques to effectively handle the nonlinear and complex motion patterns of vehicles. By employing an encoder-decoder architecture motion model, our method aims to design a more nuanced representation of vehicular dynamics.
\noindent To summarize, our contributions are the following:
\begin{itemize}
\vspace{-0.2cm}
     \item  We introduce a novel Bi-directional KF strategy for offline vehicle pose smoothing, which employs both forward and backward processing, allowing an image-based pose estimator to process video.
    \item We propose a learnable motion model integrated into our Kalman filter, thereby enabling the network to learn more complex and non-linear vehicle motion.
    \item Comprehensive experimental validation demonstrates that the proposed method outperforms the existing image-based techniques by adding temporal consistency, leading to robustness against occlusion and distant vehicles.
\end{itemize}
\begin{figure}
    \centering
    \includegraphics[trim={0cm 0.0cm 0cm 0.0cm}, clip, width=0.7\linewidth]{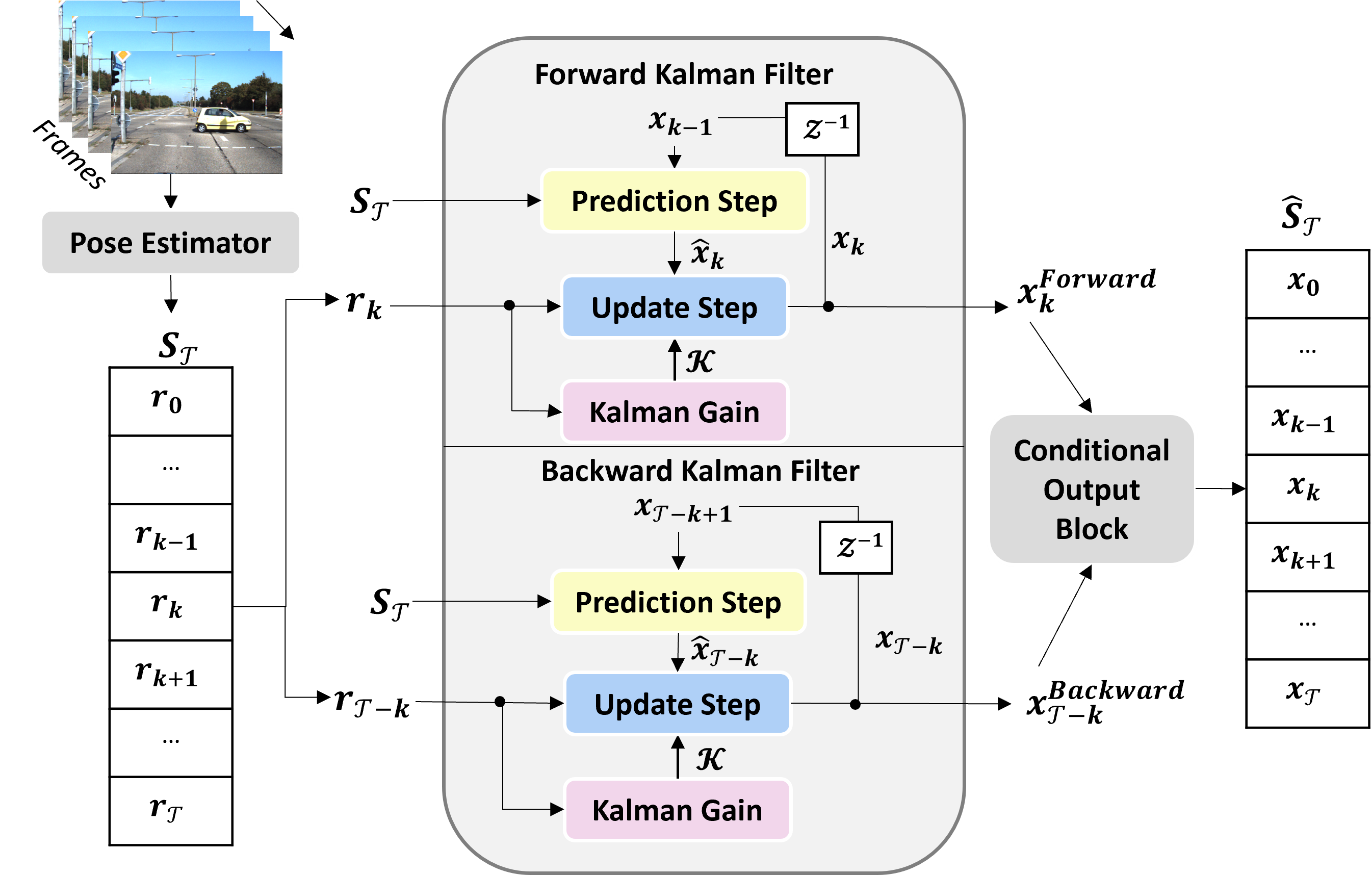}
    \caption{Schematic Overview of \textit{DeepKalPose}.}
    \label{fig:full_arch}
\vspace{-0.6cm}
\end{figure}
\vspace{-0.2cm}
\section{{Notations}}
\textit{DeepKalPose} estimates and adjusts vehicle pose estimates over sequential frames given the predictions of an existing pose estimator. Given the importance of the yaw angle in vehicle pose estimation, this letter focuses exclusively on the yaw angle within the rotation components. Our dataset, denoted as $\mathcal{D}$, comprises a series of samples ${S_{i,\mathcal{T}}}$, each representing a time-series of vehicle poses. ${S_{i,\mathcal{T}}}$ can be formally expressed as:
\begin{equation}
    S_{i,\mathcal{T}} = \{\boldsymbol{r}_{i,k}\}^{\mathcal{T}}_{k=1}
\end{equation}
Here, $\mathcal{D} =  \{S_{i,\mathcal{T}} \}^{\mathcal{N}}_{i=1}$ where $i \in \{1, ..., \mathcal{N}\}$ with $\mathcal{N}$ representing the total number of samples. The variable $k \in \{1, ... ,\mathcal{T}\}$ denotes the discrete time steps with $\mathcal{T}$ being a fixed time length of the time-series $S_{i,\mathcal{T}}$. A vehicle pose is represented as $\boldsymbol{r}_{i,k} = [x_{i,k}, y_{i,k}, z_{i,k},  \theta_{i,k}] \in \mathbb{R}^4$ where $x_{i,k},y_{i,k},z_{i,k}$ are the 3D translation components of the $i^{th}$ vehicle sample at timestep $k$ and $\theta_{i,k}$ denotes the yaw or heading angle component. 

\section{Proposed Method}
An overview of the methodology is illustrated in Figure \ref{fig:full_arch}. This method will refine the pose predictions of an existing pose estimator to produce a more accurate and temporally consistent pose time-series output $\hat{S}_\mathcal{T}$ using KF algorithm; Refer to \cite{welch1995introduction} for a comprehensive explanation of the Kalman filter algorithm. To achieve this, the filter takes the inconsistent pose prediction sequence $S_\mathcal{T}$ as input. At each iteration $k$, the measurement vector is updated as $\boldsymbol{r}_k = [\bar{x}_k, \bar{y}_k, \bar{z}_k, \bar{\theta}_k] \in S_{\mathcal{T}}$. For clarity, the symbol \space $\bar{}$ \space is used to denote measurements belonging to $S_{\mathcal{T}}$, differentiating them from the state vector used in the KF. Given the objective of tracking and correcting pose measurements, the state vector is defined as $\boldsymbol{x}_k = [x_k, y_k, z_k, \theta_k]$, thereby directing the KF to track the vehicle's pose. However, existing model-based KF methods \cite{https://doi.org/10.1049/el.2020.0374, inbook, 9626850, wu_improved_2022} have difficulties solving the temporal consistency since most time-series data from the autonomous driving dataset typically depict vehicles approaching or receding from the camera. Thus, the precision of pose estimation is significantly influenced by the vehicle’s position within the image. Specifically, the accuracy of the estimates is higher for vehicles positioned closer to the camera than for vehicles farther away. This variability in accuracy impacts the effectiveness of KF tracking as the initial measurement and the corresponding precision play a crucial role in setting the starting point of the KF. 

To address this challenge, we propose a bi-branch Kalman Filter approach. The first branch, the forward KF, processes the time-series from timestep $k=0$ to $\mathcal{T}$ where $\mathcal{T}$ is the context length of the input time sequence. Conversely, the backward KF branch processes the time-series in reverse, from the timestep $k=\mathcal{T}$  to ${0}$. By selecting, through the conditional output block (COB), the most favorable samples from the time-series—closer to the camera at the initial step—we enhance the model's ability to accurately predict and track vehicle movements. The algorithm is outlined in Algorithm \ref{alg:DeepKalPose}. We note that the proposed method acts as an offline smoother and processes the input data on chunks of $\mathcal{T}$ frames. While this approach does not render the method online, selecting sufficiently small $\mathcal{T}$ can enable near real-time processing. This method capitalizes on the higher quality of pose estimations near the camera, regardless of their position within the time-series. By feeding each time-series through both the forward and the backward EDLKF, our methodology aims to enhance the robustness and accuracy of pose estimations for distant and partially observed vehicles, as both scenarios present challenges for baseline estimations.
\begin{figure*}
    \centering
        \includegraphics[trim={0cm 0.1cm 0cm 0.3cm}, clip, width=0.8\linewidth]{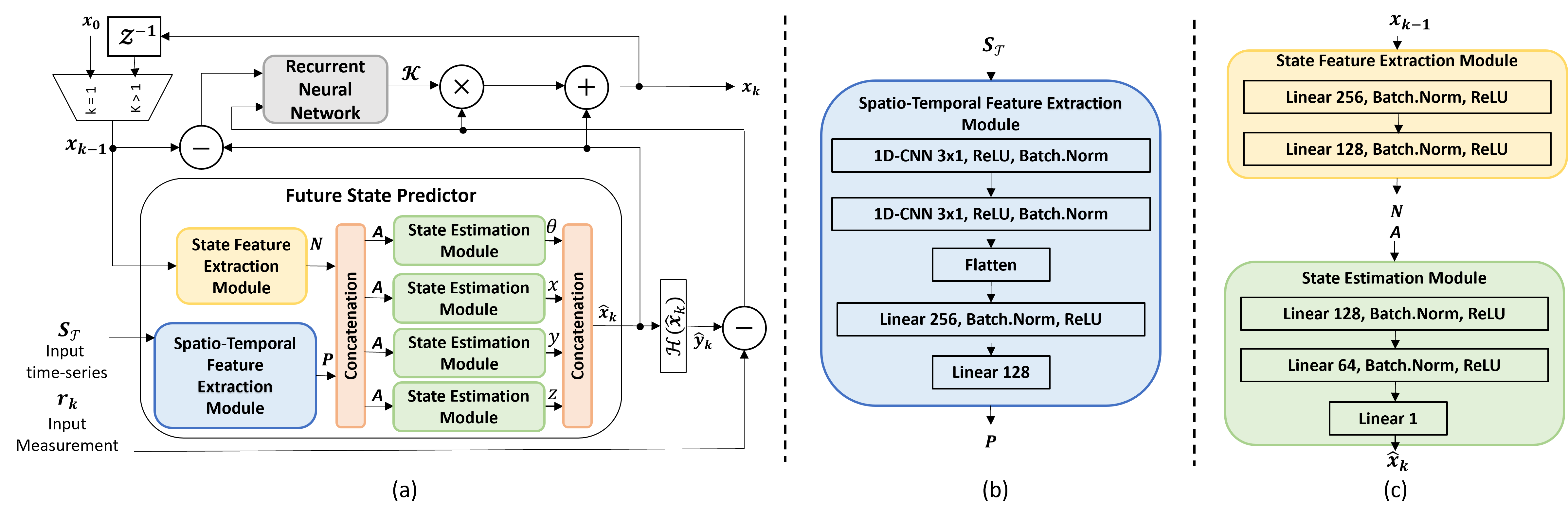}
    \vspace{-0.3cm}
    \caption{(a) Schematic of the proposed EDLKF Architecture. (b) The Spatio-Temporal Feature Extraction Module. (c) Details of the State Feature Extraction Module (Top) and State Estimation Module (Bottom). $\mathcal{Z}^{-1}$ is a unit delay, $\otimes$ is multiplication operation, $\oplus$ is a sum operation and $\ominus$ is a substraction operation.} 
\label{fig:enhanced-kalman-filter}
\vspace{-0.65cm}
\end{figure*}
\vspace{-0.2cm}
\begin{algorithm}
\caption{Overview of DeepKalPose}
\label{alg:DeepKalPose}
\begin{algorithmic}[1]
\item[] \textbf{Input:} Complete trajectory data $\mathbf{T}$ of a vehicle
\item[] \textbf{Output:} Corrected trajectory data $\mathbf{\hat{T}_{sub}} =  \{\hat{S}_{i,\mathcal{T}} \}^{\mathcal{N}}_{i=1}$
\STATE \textbf{Initialization:} Segment $\mathbf{T}$ into $\mathcal{N}$ sub-sequences $S_{i,\mathcal{T}}$ with fixed length $\mathcal{T}$ using stride one:  $\mathbf{T_{sub}} =  \{S_{i,\mathcal{T}} \}^{\mathcal{N}}_{i=1}$, where $i \in \{1, \ldots, \mathcal{N}\}$
\FOR{$i = 1$ to $\mathcal{N}$}
    \item[] \textbf{Data Pre-processing:}
    \STATE Initialize forward sequence $S^{f}_{i,\mathcal{T}} = [\mathbf{r}_{i,1}, \ldots, \mathbf{r}_{i,\mathcal{T}}]$, where $\mathbf{r}_{i,j}$ is the mean value of the valid poses if it is mis-detected
    \STATE Initialize backward sequence similar to \textit{step 3}, $S^{b}_{i, \mathcal{T}} = [\mathbf{r}_{i,\mathcal{T}}, \ldots, \mathbf{r}_{i,1}]$
    \STATE Set initial states $\boldsymbol{x}^{f} = \boldsymbol{x}^{b} = \mathbf{0}$
    \item[] \textbf{Inference:}
    \FOR{$k = 0$ to $\mathcal{T}$}
        \STATE $\boldsymbol{x}^f[k]$ =$ \text{EDLKF}^f$($S^{f}_{i,\mathcal{T}}[k]$, $S^{f}_{i,\mathcal{T}}$)
        \STATE $\boldsymbol{x}^b[k]$ = $\text{EDLKF}^b$($S^{b}_{i,\mathcal{T}}[k]$, $S^{b}_{i,\mathcal{T}}$)
    \ENDFOR
    \STATE $\hat{S}_{i,\mathcal{T}} = \text{COB}[\boldsymbol{x}^{f}, \boldsymbol{x}^{b}]$
\ENDFOR

\end{algorithmic}
\end{algorithm}
\vspace{-0.2cm}
The Kalman Filter necessitates a thorough understanding of the system’s dynamics and noise characteristics to function effectively. However, for the pose estimation task, our limited knowledge of the vehicle’s systems presents a significant challenge. Therefore, instead of using a traditional model-based Kalman Filter, inspired by KalmanNet \cite{9733186}, our \textit{DeepKalPose} replaces the computation the Kalman gain $\mathcal{K}$ and both of the covariances $\boldsymbol{v}_{k}$ and $\boldsymbol{w}_{k}$ by a Recurrent Neural Network (RNN). Moreover, we propose a novel module, named Future State Predictor (FSP), which is able to learn a predictive motion model, as illustrated in Figure \ref{fig:enhanced-kalman-filter}.  The FSP block follows an encoder-decoder architecture where the encoder takes as input the state  $\boldsymbol{x}_{k-1}$ and the sequence that is being processed $S_{\mathcal{T}}$ with $\mathcal{T}$ the context length of the sequence and $k \in \{0, ... ,\mathcal{T}\}$ representing the timestep of the KF iterative process. More precisely, the top branch of the encoder, namely the State Feature Extraction Module (SFEM), processes the current state $\boldsymbol{x}_{k-1}$ to derive a feature vector $\mathbf{N}$. The SFEM captures the relevant aspects of the current state that may influence the future state. The lower branch of the encoder, namely Spatio-Temporal Feature Extraction Module (STFEM) is dedicated to extracting features $\mathbf{P}$ from the entire sequence $S_{\mathcal{T}}$. This module analyzes local spatial and temporal patterns in the sequence of 3D position vectors, aiding in predicting the next state. Spatial patterns show the relationships among features like translation and rotation at each timestep, while temporal patterns track changes over time. In contrast, the SFEM focuses on immediate characteristics of the current state. The feature vectors from the SFEM and the STFEM are concatenated into a combined feature vector $\mathbf{A}$.
The decoder consists of four State Estimation Modules (SEM) which will transform individually the feature vector $\mathbf{A}$ into the three translation components and the rotation component of the predicted state vector $\boldsymbol{\hat{x}}_{k}$.
To train this model, we have used an L1-loss for the translation components as follows :
$\mathcal{L}_\alpha = \lvert \hat{\alpha} - \alpha \rvert$ where $\alpha \in \{x, y, z\}$ and $\hat{\alpha}$ denotes the predicted translation values. The rotation loss is defined as: $\mathcal{L}_\theta = 1 - \text{cos}(\hat{\theta}, \theta)$ where $\text{cos}$ is the cosine similarity between the predicted heading angle $\hat{\theta}$ and the ground truth $\theta$. The loss per mini-batch $\mathcal{B}$ with the mini-batch size $M$ < $\mathcal{N}$ is denoted as : 
$
   \mathcal{L}_\mathcal{B} = \frac{1}{M}\sum_{j=1}^{M} \frac{1}{\mathcal{T}}\sum_{k=0}^{\mathcal{T}} (\mathcal{L}_x + \mathcal{L}_y+ \mathcal{L}_z+ \mathcal{L}_\theta)_{j,k}
   \label{eq:aed}
$

\section{Experimental setup}
The KITTI RAW dataset \cite{geiger2012we} is a widely used dataset for monocular object pose estimation {\cite{zhang2021objects, peng2022did, inproceedings} and tracking \cite{cao2203observation, kim2022polarmot, wu20213d}}. The dataset comprises 51 videos, divided into 39 for training and 12 for validation. It includes a total of 9903 images for training and 2977 images for validation. We segment each sequence into fixed length with 20 timesteps using a stride of 1 for continuity. In total, we have 674 vehicle trajectory patches for training and 375 for validation. To train  \textit{DeepKalPose}, we extract the pose predictions from an existing vehicle pose estimator \cite{inproceedings, lyu2022mono6d} as $\boldsymbol{r}_k$. To handle missing data from a non-detected vehicle by the vehicle pose estimator, we applied mean substitution. The network is optimized by Adam Optimizer \cite{kingma2017adam} with a learning rate of 0.001 and a weight decay of 0.00001. We take a batch size of 128 on 1 Nvidia GeForce RTX 2080 (12G). The iteration number for the training process is set to 4,000. Evaluation metrics used to compare against D4LCN \cite{inproceedings} include 3D precision-recall curves with a 3D bounding box IoU threshold of 0.7 and 0.5 for cars as this method performs an object detection step. Against Mono6D \cite{lyu2022mono6d}, evaluation is performed using, for translation, the Average Relative Euclidean Distance (\textbf{ARED}). For rotation, we use the accuracy with threshold $\delta$, denoted \textbf{Acc}$(\delta)$, and the median error, \textbf{Mederr}, in degrees. Following the evaluation of D4LCN in \cite{inproceedings}, we only consider the detected vehicle with a 2D bounding box IoU threshold of 0.5.

\section{Experimental results}
 In Table \ref{tab:comparison_D4LCN}, we compared our proposed method, \textit{DeepKalPose}, with the state-of-the-art (SOTA) method D4LCN \cite{inproceedings}.  One can note that with \textit{DeepKalPose}, we can significantly outperform D4LCN \cite{inproceedings}. The average precision @70 of D4LCN improves to 31.12\% for Easy, 24.82\% for Moderate, and 16.70\% for Hard scenarios, compared to 28.07\%, 21.56\%, and 14.13\% respectively without \textit{DeepKalPose}. The same behavior is observed for the AP@50.
 
 \begin{figure*}
    \centering
    \includegraphics[trim={0cm 0cm 0cm 0cm}, clip, scale=0.38]{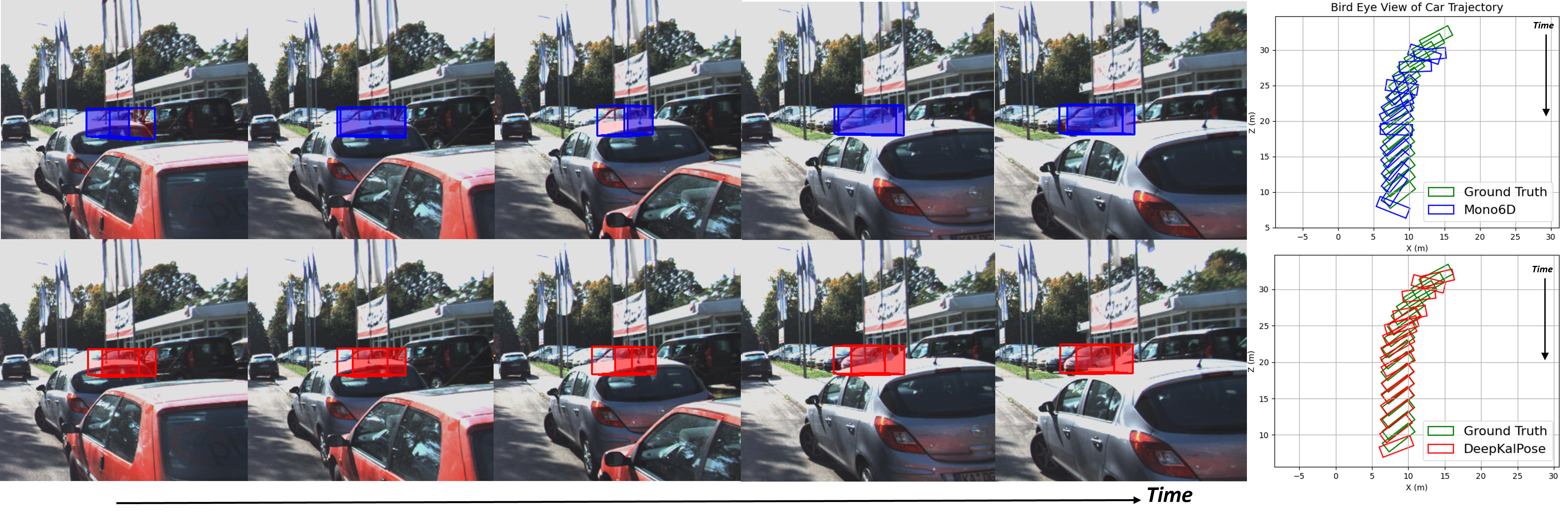}
    \vspace{-0.3cm}
    \caption{Qualitative results demonstrating improved car trajectory estimation on an occluded and distant vehicle. The sequence displays the projected 3D bounding boxes: the upper row illustrates results from Mono6D \cite{lyu2022mono6d}, while the lower row illustrates results from our proposed methodology. The two figures on the extreme right detail the temporal Bird's Eye View (BEV) comparison between the 3D BB estimations by Mono6D and \textit{DeepKalPose}, respectively.}
    \label{fig:qual_results}
    \vspace{-0.5cm}
\end{figure*}

\begin{table}[!h]
\caption{Method comparison in terms of average precision. The methods denoted with * were retrained.}
\label{tab:comparison_D4LCN}
\centering
\resizebox{0.9\columnwidth}{!}{%
\begin{tabular}{lcccccc}
\hline
\textbf{Method} & \multicolumn{3}{c}{\textbf{Average Precision @70}}  & \multicolumn{3}{c}{\textbf{Average Precision @50} }\\ \hline
                & \textbf{Easy} & \textbf{Moderate} & \textbf{Hard}  & \textbf{Easy} & \textbf{Moderate} & \textbf{Hard} \\ \hline
D4LCN* \cite{inproceedings}           & 28.07            & 21.56              & 14.13    & 67.08            & 52.55              & 35.74          \\ \hline
Our Method & \textbf{31.12}      & \textbf{24.82}            & \textbf{16.70}  & \textbf{68.67}      & \textbf{53.74}            & \textbf{36.89}  \\ \hline
\end{tabular}}
\vspace{-0.3cm}
\end{table}
 
In Table \ref{tab:comparison_mono6D}, the results illustrate the performance differences among the original Mono6D, Mono6D combined with a model-based Kalman filter (MB-KF), and Mono6D enhanced with DeepKalPose. The design of the MB-KF is inspired by the state-of-the-art tracking methodology presented in \cite{wu20213d}. In the table, the best results are in bold and the second-best are underlined. One can observe that the integration of \textit{DeepKalPose} led to notable improvements in vehicle 6D pose estimation. As such, the \textbf{ARED} sees an improvement from 5.34\% to 3.90\% while the MB-KF barely decreases the error. In addition, our method improves the heading angle orientation accuracy (\textbf{Acc($\delta$)}) by 4.52\% for $\delta = \frac{\pi}{6}$ and 2.08\% for $\delta=\frac{\pi}{18}$ although \textbf{Mederr} slightly increases, where the traditional method tends to replicate the results from the baseline. This result suggests that the EDL Kalman filter's smoothing operation leads to more stable rotations but at the expense of smaller, more frequent errors.
 \begin{table}
\caption{Comparison with the existing methods. The methods denoted with * were retrained.}
\label{tab:comparison_mono6D}
\centering
\resizebox{0.9\columnwidth}{!}{%
\begin{tabular}{lcccccc} 
\hline
 \textbf{Method} & \textbf{ARED} \boldmath$\downarrow$ & \textbf{Acc ($\frac{\pi}{6}$)} \boldmath$\uparrow$  & \textbf{Acc ($\frac{\pi}{18}$)} \boldmath$\uparrow$  & \textbf{Mederr} \boldmath$\downarrow$\\ \hline
 
 Mono6D* \cite{lyu2022mono6d}                & 5.34\%       & {84.66\%}        & \underline{{65.41\%}} & \textbf{4.94$^{\circ}$}   \\ 
 Mono6D* + MB-KF \cite{wu20213d}                & \underline{4.92\%}       & \underline{{84.69\%}}        & {65.34\%} & \underline{{5.03$^{\circ}$}}   \\ 
Mono6D + {DeepKalPose} (ours) & \textbf{ 3.90\% }       & \textbf{89.18\%}        & \textbf{67.47\%} & 5.46$^{\circ}$    \\ \hline
 
\end{tabular}}
\vspace{-0.3cm}
\end{table}
 \begin{table}
\caption{Evaluation results at different occlusion levels.}
\label{tab:occlusion}
\centering
\resizebox{0.9\columnwidth}{!}{%
\begin{tabular}{lcccccc} 
\hline
 \textbf{Method} & Occlusion Level & \textbf{ARED} \boldmath$\downarrow$ & \textbf{Acc ($\frac{\pi}{6}$)} \boldmath$\uparrow$  & \textbf{Acc ($\frac{\pi}{18}$)} \boldmath$\uparrow$  & \textbf{Mederr} \boldmath$\downarrow$\\ \hline
 
  Mono6D \cite{lyu2022mono6d}  &\textit{Visible}              & 4.06\%     & {92.35\%}        & {77.40\%} & \textbf{3.19$^{\circ}$}  \\  
   &\textit{Fully Occluded}              & 6.97\%       & {73.86\%}        & {49.07\%} & 10.19$^{\circ}$  \\ \hline
  Our Method &\textit{Visible}    & \textbf{{ 3.64\% }}       & \textbf{93.84\%}        & \textbf{79.00\%} & {4.21}$^{\circ}$ \\
 
 &\textit{Fully Occluded}  & \textbf{ 4.21\% }       & \textbf{82.50\%}        & \textbf{51.53\%} & \textbf{9.30}$^{\circ}$
\\ \hline
\end{tabular}}
\vspace{-0.3cm}
\end{table}

\begin{figure}
    \centering
    \includegraphics[width=0.73\linewidth]{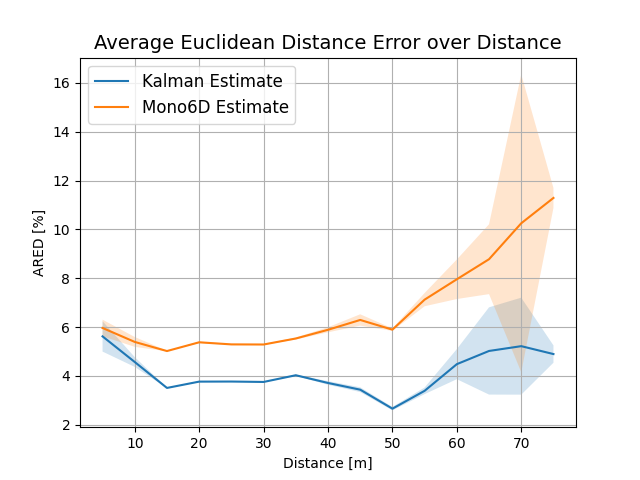}
        \vspace{-0.4cm}
    \caption{The ARED of Mono6D and proposed method against distance. Solid lines represent the mean values, while the shaded areas indicate the variance. }
    \label{fig:distant_vehicles}
\vspace{-0.75cm}
\end{figure}

\begin{table}
\caption{Evaluation results at different depths.}
\label{tab:depth}
\centering
\resizebox{0.9\columnwidth}{!}{%
\begin{tabular}{lcccccc} 
\hline
 \textbf{Method} & Depth & \textbf{ARED} \boldmath$\downarrow$ & \textbf{Acc ($\frac{\pi}{6}$)} \boldmath$\uparrow$  & \textbf{Acc ($\frac{\pi}{18}$)} \boldmath$\uparrow$  & \textbf{Mederr} \boldmath$\downarrow$\\ \hline
 
  Mono6D \cite{lyu2022mono6d}  
    & 0m-40m         & 5.18\%       & {84.96\%}        & {65.86\%} & \textbf{4.85$^{\circ}$}   \\ 
   & 40m-$\infty$           & 6.28\%       & {82.94\%}        & {\textbf{62.81}\%} & \textbf{5.51$^{\circ}$}   \\ \hline
  Our Method& 0m-40m & \textbf{3.96\% }       & \textbf{89.53\%}        & \textbf{68.29\%} & 5.24$^{\circ}$    \\
 & 40m-$\infty$ & \textbf{ 3.55\% }       & \textbf{87.12\%}        & {62.67\%} & 7.02$^{\circ}$    \\ \hline
 
\end{tabular}}
\vspace{-0.7cm}
\end{table}
Furthermore, in scenarios of partial or complete vehicle occlusion in Table \ref{tab:occlusion}, this experiment reveals that \textit{DeepKalPose} is robust against occlusion demonstrating a reduction of ARED error from 6.97\% to 4.21\%. Thanks to the predictive capabilities of the KF and the use of the temporal information from past measurements, \textit{DeepKalPose} can adjust the initial noisy measurements from the pose estimator due to occlusion and put more emphasis on the learned predicted state vector. Moreover, in Table \ref{tab:depth}, we evaluate both methods for various vehicle distances from the camera. Specifically, for distances exceeding 40 meters (considered as far-away-object detection), our proposed method demonstrates a notable decrease in ARED from 6.28\% to 3.55\%, indicating a 2.73\% improvement. One can note that \textit{DeepKalPose} utilizes past detections and temporal information effectively to mitigate the impact of distance on model performance, in contrast to the standard pose estimator whose performance diminishes with increasing distance. Figure \ref{fig:distant_vehicles} confirms the efficacy of \textit{DeepKalPose} against far-vehicles pose estimation as illustrated by a bigger gap in ARED between the two methods when the distance is increasing. In Figure \ref{fig:qual_results}, we introduce a comparative visualization on a far-away and occluded vehicle highlighting the flickering artifacts present in the Mono6D estimates, where the 3D bounding boxes show significant rotational fluctuations over time. In contrast, our method introduces a temporal component that significantly enhances stability, with poses temporally coherent and reduced visual artifacts.

In Table \ref{tab:abl_bikalman}, we mainly study the impact of the STFEM, and the impact of both the forward branch (\textbf{$\mathbf{B_{forw}}$}) and backward branch (\textbf{$\mathbf{B_{back}}$}). One can observe that for better translation performance, both directions should be taken into account. On top of that, Table \ref{tab:abl_bikalman} highlights the importance of the STFEM in the Kalman filtering process. A notable limitation of \textit{DeepKalPose} is its operation as an offline algorithm, processing and analyzing the video data after it is fully processed by the pose estimator, which constrains its online applicability. While the method provides high accuracy and consistency in vehicle pose estimation, it's not ideal for scenarios where a frame needs to be processed immediately once captured. Future work will address this problem, moving from an offline method to an online method. Furthermore, for further improvements, we plan to replace the current empirical Conditional Output Block (COB) with a deep-learning-based confidence network.

\begin{table}
\caption{Ablation studies on the KITTI validation set.}
\label{tab:abl_bikalman}
\centering
\resizebox{0.9\columnwidth}{!}{%
\begin{tabular}{cccccccc} 
\hline
  \textbf{$\mathbf{STFEM}$} & \textbf{$\mathbf{B_{forw}}$} & \textbf{$\mathbf{B_{back}}$} & \textbf{ARED} \boldmath$\downarrow$ & \textbf{Acc ($\frac{\pi}{6}$)} \boldmath$\uparrow$  & \textbf{Mederr} \boldmath$\downarrow$\\ \hline
 
  \Checkmark & \Checkmark &  & 4.26\% & \underline{88.78}\%   & \textbf{5.10}$^{\circ}$   \\ 
 & \Checkmark  & \Checkmark & \underline{4.14\% }  & {84.79\%} & 5.68$^{\circ}$\\
   \Checkmark  & \Checkmark & \Checkmark  & \textbf{ 3.90\% }   & \textbf{89.18\%}   & \underline{5.46$^{\circ}$} \\ \hline
\end{tabular}}
\vspace{-0.4cm}
\end{table}

\section{Conclusion}
In this letter, we present DeepKalPose, an innovative approach integrating a deep learning-based Kalman Filter to enhance temporal consistency in monocular vehicle 6D pose estimation for video data. Leveraging a learnable motion model, DeepKalPose effectively captures the complex, nonlinear motion patterns of vehicles, significantly surpassing existing methods in accuracy and consistency for 4D object detection and 6D pose estimation tasks. The experimental results demonstrate the model's effectiveness in improving pose estimation accuracy and consistency from single-view images, particularly in challenging conditions such as far-object detection and occlusion. These results affirm the efficacy of our deep-learning-based Kalman Filter in video-based pose estimation and suggest its potential to enhance intelligent transport systems.

\begin{acks}
This work is funded by Innoviris within the research project TORRES.\\
\textit{Conflict of interest statement}: The authors declare no conflict of interest.\\
\textit{Data availability statement}: The data that support the findings of this study are available in KITTI RAW
Dataset at [https://www.cvlibs.net/datasets/kitti/], reference number \cite{geiger2012we}.\\
\textit{Credit contribution statement:}
\begin{enumerate}
    \item Leandro Di Bella: Conceptualization – Data Curation – Formal Analysis – Investigation - Methodology – Software - Validation – Visualization – Writing-Original Draft – Writing-Review \& Editing
    \item Yangxintong Lyu: Conceptualization - Methodology – Validation - Project Administration – Supervision - Writing-Review \& Editing
    \item Adrian Munteanu: Conceptualization - Project Administration – Supervision - Writing- Review \& Editing – Funding Acquisition
\end{enumerate}
\end{acks}


\bibliographystyle{ieeetr}
\bibliography{iet-ell.bib}

\newpage

\end{document}